%% file: ARLET_2024.tex
\documentclass{article}

\usepackage[preprint]{ARLET_2024}

\usepackage{algorithmic,algorithm}
\usepackage[utf8]{inputenc} 
\usepackage[T1]{fontenc}    
\usepackage{hyperref}       
\usepackage{url}            
\usepackage{booktabs}       
\usepackage{amsfonts}       
\usepackage{nicefrac}       
\usepackage{microtype}      
\usepackage{xcolor}         
\usepackage{subcaption}
\usepackage{listings}
\usepackage{wrapfig}
\usepackage[export]{adjustbox}

\title{BenchMARL: Benchmarking Multi-Agent Reinforcement Learning}

\author{%
  Matteo Bettini\thanks{Work done during an internship at PyTorch, Meta.}\\
  University of Cambridge\\
  \texttt{mb2389@cl.cam.ac.uk} \\
  \And
  Amanda Prorok \\
  University of Cambridge \\
  \texttt{asp45@cl.cam.ac.uk} \\
  \And
  Vincent Moens \\
  PyTorch, Meta \\
  \texttt{vmoens@meta.com} \\
}

\begin{document}

\maketitle

\begin{abstract}
The field of Multi-Agent Reinforcement Learning (MARL) is currently facing 
a reproducibility crisis. While solutions for standardized reporting have been proposed to address the issue, we still lack a benchmarking tool that enables standardization and reproducibility, while leveraging cutting-edge Reinforcement Learning (RL) implementations. 
In this paper, we introduce BenchMARL, the first MARL training library created to enable standardized benchmarking across different algorithms, models, and environments. 
BenchMARL uses TorchRL as its backend, granting it high performance and maintained state-of-the-art implementations while addressing the broad community of MARL PyTorch users. 
Its design enables systematic configuration and reporting, thus allowing users to create and run complex benchmarks from simple one-line inputs. BenchMARL is open-sourced on GitHub at: \url{https://github.com/facebookresearch/BenchMARL}.
\end{abstract}

\begin{figure}[h]
    \centering
    \includegraphics[width=0.6\linewidth]{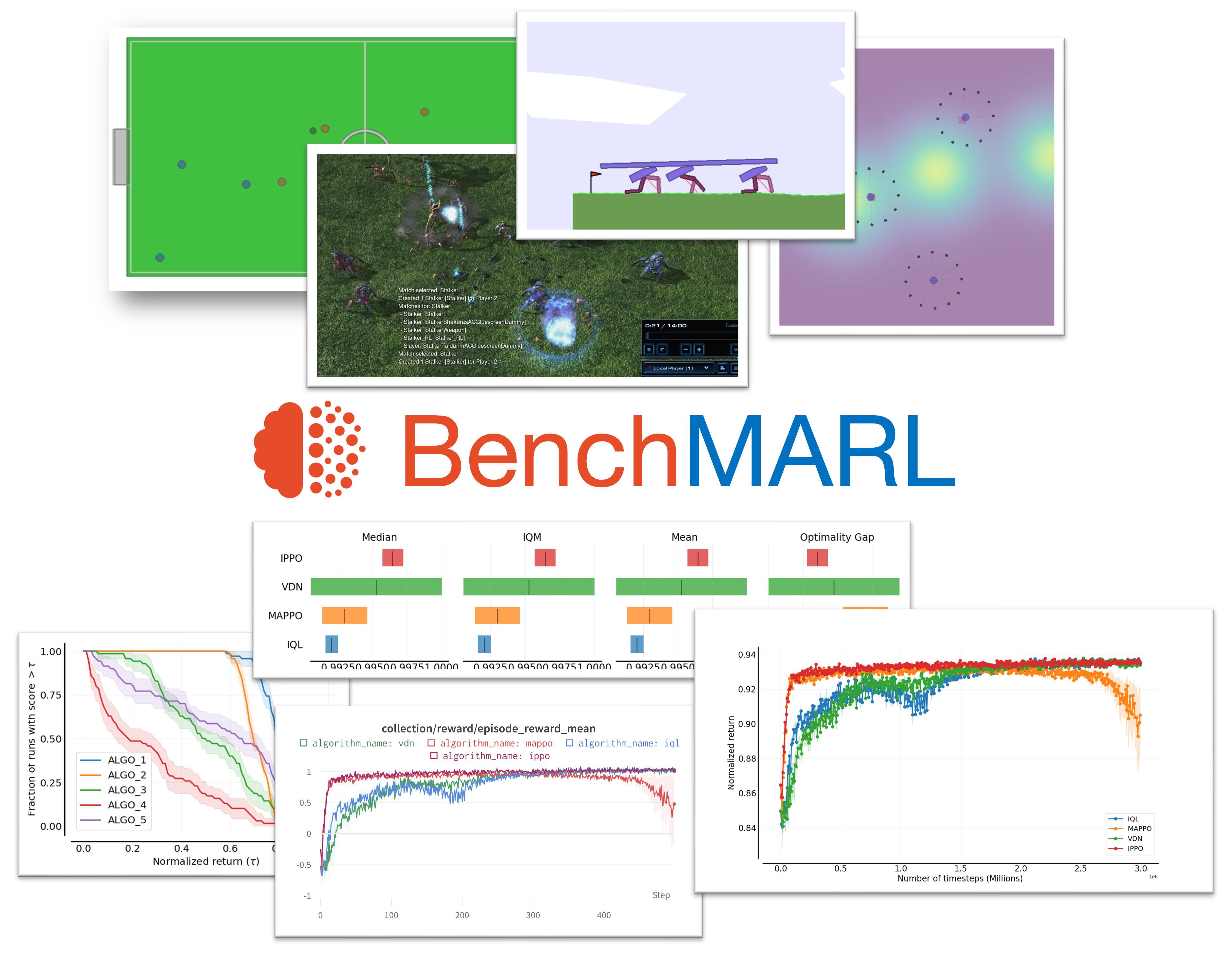}
    \caption{BenchMARL enables comparisons across different Multi-Agent Reinforcement Learning (MARL) algorithms, models, and tasks while focusing on standardization and reproducibility.}
    \label{fig:benchmarl_hero}
\end{figure}

\section{Introduction}

\input{sections/introduction}

\section{Related work}
\input{sections/realted}

\section{BenchMARL}

\input{sections/benchmarl}

\subsection{Components}
\input{sections/components}

\subsection{Features}
\input{sections/features}

\section{Experiments}
\input{sections/experiments}

\section{Conclusion}
\input{sections/conclusion}

\section*{Acknowledgments}
The majority of the work was done during an internship at PyTorch, Meta.
This work was supported in part by European Research Council (ERC) Project 949940 (gAIa) and ARL DCIST CRA W911NF-17-2-0181. 

\bibliographystyle{unsrtnat}
\bibliography{bibliography}




\end{document}

%% file: sections/introduction.tex
The Multi-Agent Reinforcement Learning (MARL) community in PyTorch is evergrowing.
Despite this, there exists a persistent fragmentation of experimental tools and standards in the field.
The PyTorch-backed TorchRL project~\citep{bou2023torchrl} successfully addressed this issue in the broader Reinforcement Learning (RL) domain, 
providing a state-of-the-art library adopted by thousands of users.
In this work, we introduce BenchMARL: a MARL training library that leverages TorchRL as its backend while focusing on standardization of MARL experiments.

Thanks to the flexibility of the TorchRL backend, BenchMARL enables reproducibility and benchmarking across different MARL algorithms, models, and environments. Its mission is to present a standardized interface that allows easy integration of new algorithms and environments to provide a fair and systematic comparison with existing solutions. 
Its core design tenets are: (1) \textit{Reproducibility}, achieved via standardization of configuration using the Hydra framework~\citep{Yadan2019Hydra};
(2) \textit{Standardized plotting and reporting}, achieved by integrating with the statistically-rigorous tools proposed by~\cite{gorsane2022towards}
\footnote{Based on the NeurIPS 21 Outstanding Paper by~\cite{agarwal2021deep}\label{foot:marl-eval}}; 
(3) \textit{TorchRL backend}, which grants high performance and state-of-the-art RL implementations; (4) \textit{Experiment independence}, achieved via an experiment class that is agnostic to the choices of algorithm, model, or task; (5) \textit{Easy integration of new solutions}, achieved via simple abstract interfaces.

This set of tenets allows BenchMARL users to go from simple command line inputs to statistically rigorous output reports (see \autoref{fig:benchmarl}) without extensive experimental domain knowledge. In doing so, the library uses (when possible) solutions from the single-agent RL domain, avoiding custom re-implementations and leveraging the extensive benchmarking already performed on TorchRL algorithms~\citep{bou2023torchrl}. High-performance is guaranteed thanks to several features such as: vectorized simulation~\citep{bettini2022vmas}, vectorized training (using \lstinline[columns=fixed]{torch.vmap}) over the agents' parameters and data, and a SLURM~\citep{yoo2003slurm} launcher backend for running experiments on High Performance Computing (HPC) clusters.

In the following, we introduce the library, providing a overview of its structure and discussing its components (\autoref{sec:components}) and features (\autoref{sec:features}). We then report results from an initial experimental comparison, consisting of benchmarking the 9 algorithms available in BenchMARL on 3 VMAS~\citep{bettini2022vmas} tasks representing high-level multi-robot control problems (\autoref{sec:experiments}). 
With this effort, we hope to stimulate more researchers to adopt shared standards for configuration and reporting towards a standardization of MARL experimental pipelines.

%% file: sections/realted.tex
The recent popularity of MARL has exacerbated the fragmentation of shared community standards and tools, with new libraries being frequently introduced, each one focusing on specific algorithms, environments, or models.
Popular examples are PyMARL~\citep{samvelyan19smac} and its extensions: PyMARL2~\citep{hu2021rethinking} and EPyMARL~\citep{papoudakis2021benchmarking}, which are limited to environments with discrete action spaces.
Furthermore, these libraries often implement algorithmic components from scratch, without leveraging native and stable baselines from the single-agent RL community. 
MARLlib~\citep{hu2023marllib} addresses this problem by basing on the RLlib framework~\citep{liang2018rllib}. However, RLlib presents significant limitations due to its aim of being agnostic to the underlying learning framework, which causes it to lack core features for state-of-the-art benchmarking, such as support for vectorized environments. In contrast to these PyTorch-based libraries, BenchMARL focuses on covering the whole breadth of MARL algorithms, models, and tasks (e.g., continuous vs discrete actions and state, competitive vs cooperative, vision vs vector observation) while being compatible with state-of the art vectorized simulation.

Concurrent to our work, other MARL libraries have been proposed in the Jax~\citep{jax2018github} ecosystem~\citep{de2021mava,rutherford2024jaxmarl}.
With Jax being a promising emergent machine learning framework, these projects are important complementary tools to our work,which is in turn focused on the PyTorch domain.
To promote cross-framework comparisons, BenchMARL adopts the same reporting format as Mava~\citep{de2021mava}, making interactions easier for experimentalists from either domain.
Lastly, we note that these projects often propose single-file implementations, while BenchMARL focuses on providing abstractions and components that can be reused across experiments.

The fragmentation of the MARL domain has recently led to a reproducibility crisis, highlighted by \cite{gorsane2022towards}. While the authors propose a set of tools for results’ reporting, there is still the need for a standardized library to run such benchmarks. 
BenchMARL's mission is to provide such a benchmarking library for MARL, integrating with the reporting tools proposed and using TorchRL as an efficient, tried and tested backend.

%% file: sections/benchmarl.tex
\begin{figure}[t]
    \centering
    \includegraphics[width=\linewidth]{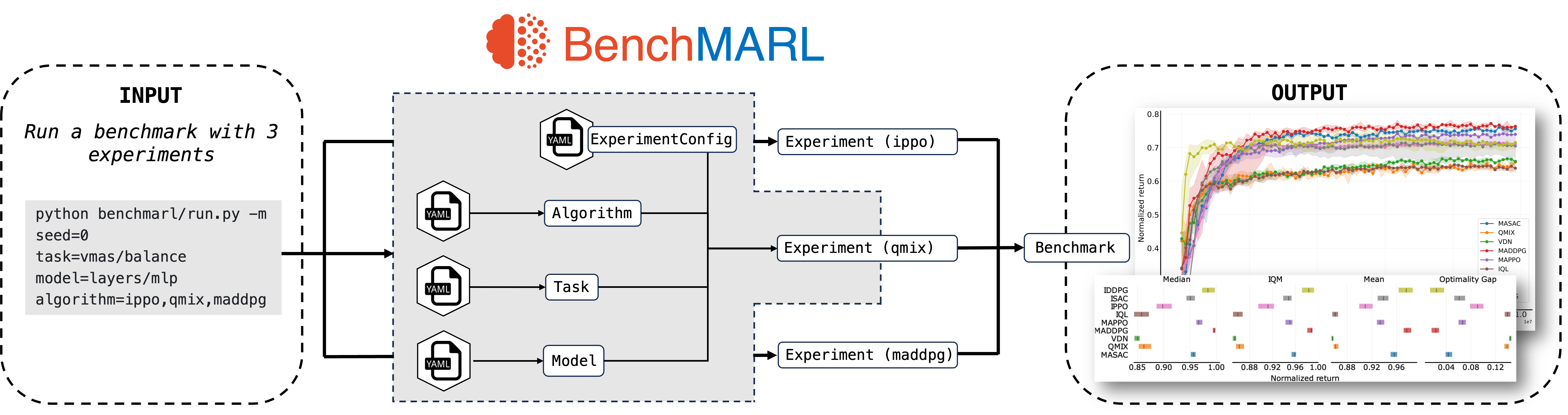}
    \caption{BenchMARL execution diagram. Users run benchmarks as sets of experiments, where each experiment loads its components from the respective YAML configuration files.}
    \label{fig:benchmarl}
\end{figure}

BenchMARL tackles its reproducibility goals via defining unifying abstractions over MARL training components. Components are gathered into experiments that are agnostic of their specific implementations. Structured configurations allow to easily run multiple experiments to create a benchmark, making it possible for users to go directly from one-line inputs to benchmarking plots. This process is depicted in \autoref{fig:benchmarl}. 
In the following, we illustrate the components and features that enable this pipeline.

%% file: sections/components.tex
\label{sec:components}

BenchMARL has a few core components, which correspond to classes in the codebase. Each component has a default YAML configuration in a dedicated directory.
We now introduce the core components of the library.

\paragraph{Experiment.} An experiment is a training run in which an algorithm, a task, and a model are fixed. Experiments are configured by passing the choices for these components alongside a seed and the experiment's hyperparameters. They cover both on-policy and off-policy algorithms, discrete and continuous actions, probabilistic and deterministic policies, and competitive and cooperative scenarios.
Experiments will try to match versions of algorithms and tasks on a best-effort basis: for example, if a task has continuous actions, and an algorithm supports both continuous and discrete actions, the experiment will instantiate the version of the algorithm compatible with the task.
To address the heterogeneity of the MARL ecosystem, experiments leverage the agent grouping mechanism available in TorchRL. This mechanism allows to define which agents should have their data stacked in a group (to benefit from vectorization) and which agents should have their data as separate entries (due to heterogeneity in their shapes).
This is a unique feature of BenchMARL that allows compatibility with both environments that stack all agent data together (e.g., SMACv2~\citep{ellis2022smacv2}) and environments that keep agent data separate (e.g., PettingZoo~\citep{terry2021pettingzoo}).
During training, each group will be trained independently and agents within a group will be able to reuse most single-agent RL components, just with the addition of the agent dimension in the data shape.
At a high-level, an experiment iteration is composed of the following steps: (1) collection: which is vectorized for compatible environments or parallel/sequential for other environments; (2) addition of data to the replay buffers of each agent group; (3) training: where, for each agent group, data is sampled from the replay buffer and optimizer steps are performed, (4) evaluation: where agents are evaluated on a test environment.
An experiment can be launched from the command line using Hydra or directly from a Python script.

\paragraph{Benchmark.} A benchmark is a collection of experiments that can vary in task, algorithm, or model. Where possible, a benchmark shares hyperparameters across its experiments. Benchmarks allow to compare different MARL components in a standardized way. A benchmark can be launched from the command line using Hydra or directly from a python script. For example, by looking at the input code in \autoref{fig:benchmarl}, we can see that it is sufficient to provide a list of components (e.g., algorithms, tasks) to instantiate a benchmark. Thanks to Hydra, it is then possible to specify custom backends for running the experiments in a benchmark, such as: sequential, in parallel processes, or in separate SLURM~\citep{yoo2003slurm} submissions on an HPC cluster.

\paragraph{Algorithms.} Algorithms are an ensemble of components (e.g., loss, replay buffer) that determine the training strategy. In \autoref{tab:algorithms} we report and classify the algorithms available in BenchMARL. 
As we can see, BenchMARL provides a wide range of algorithms, allowing our benchmarks to span over the wide range of different options available in the MARL domain.
We further provide novel implementations (MASAC, ISAC) based on the SAC~\citep{haarnoja2018soft} algorithm. These are multi-agent adaptions of the single-agent algorithm with (MASAC) and without (ISAC) full-observability in the critic.
All our algorithms have the option of sharing parameters within agent groups for actors and critics (where applicable). Custom algorithms can further be designed by combining this choice with the algorithm and the model. For example, the HetGPPO~\citep{bettini2023hetgppo} algorithm can be obtained by using IPPO without parameter sharing and with a Graph Neural Network (GNN) model for actor and critic.

\begin{table}[t]
    \centering
    \caption{Algorithms in BenchMARL.}
    \label{tab:algorithms}
    \resizebox{\linewidth}{!}{%
    \begin{tabular}{lcccccc}
    \hline
        \textbf{Name} & \textbf{On/Off policy} & \textbf{Actor-critic} & \textbf{Full-observability in critic} & \textbf{Action compatibility} & \textbf{Probabilistic actor}  \\ \hline
        MAPPO~\citep{yu2021surprising} & On & Yes & Yes & Continuous + Discrete & Yes  \\ 
        IPPO~\citep{de2020independent} & On & Yes & No & Continuous + Discrete & Yes \\ 
        MADDPG~\citep{lowe2017multi} & Off & Yes & Yes & Continuous & No  \\ 
        IDDPG & Off & Yes & No & Continuous & No \\ 
        MASAC & Off & Yes & Yes & Continuous + Discrete & Yes  \\ 
        ISAC & Off & Yes & No & Continuous + Discrete & Yes  \\ 
        QMIX~\citep{rashid2018qmix} & Off & No & NA & Discrete & No  \\ 
        VDN~\citep{sunehag2017value} & Off & No & NA & Discrete & No  \\ 
        IQL~\citep{Tan1993} & Off & No & NA & Discrete & No  \\ \hline
    \end{tabular}
    }
\end{table}

\begin{figure}[t]
    \newcommand{\subfigsize}{0.19}
     \centering
       \begin{subfigure}{\subfigsize\linewidth}
         \centering
        \includegraphics[width=\linewidth,frame]{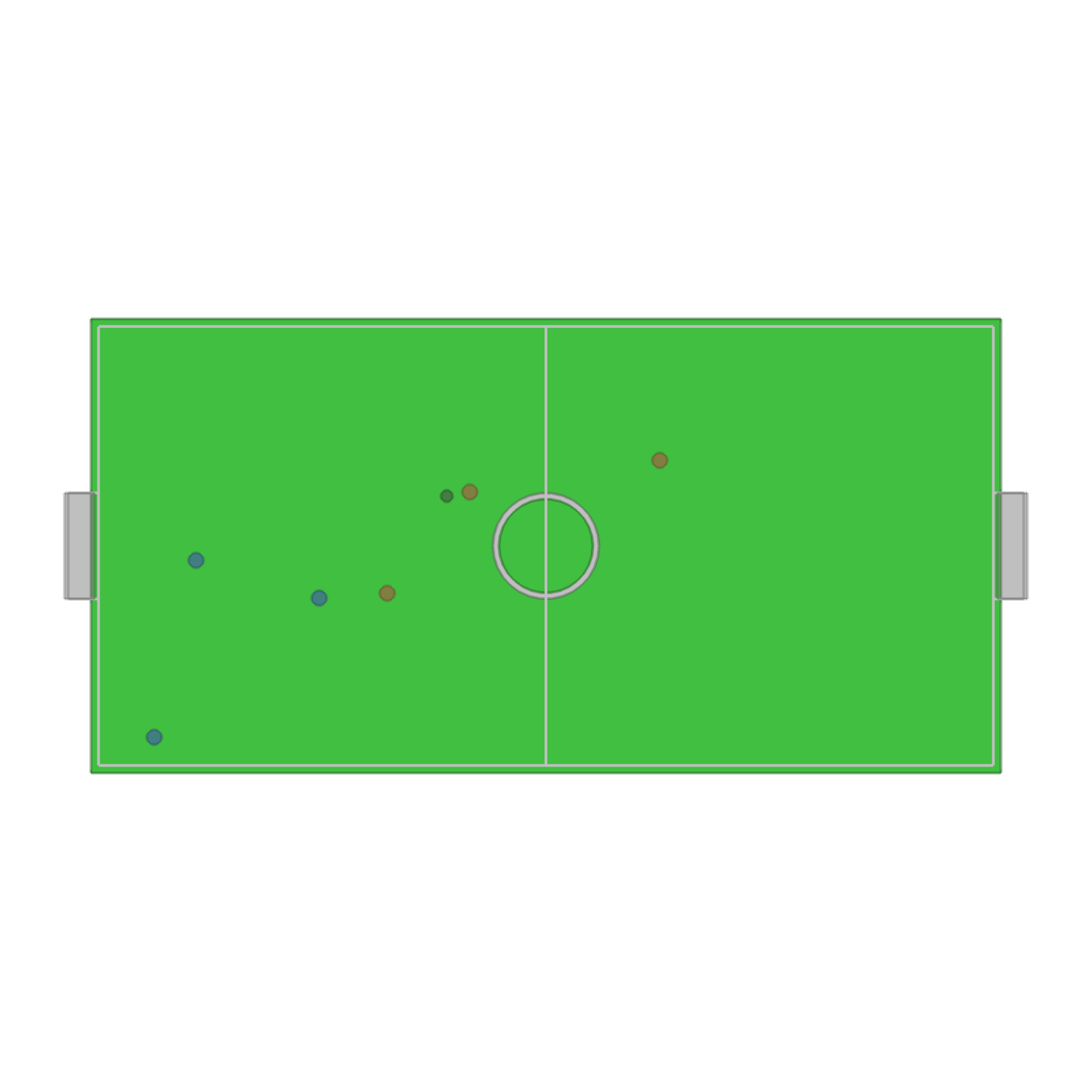}
         \caption{VMAS}
     
     \end{subfigure}%
          \begin{subfigure}{\subfigsize\linewidth}
         \centering
         \includegraphics[width=\linewidth,frame]{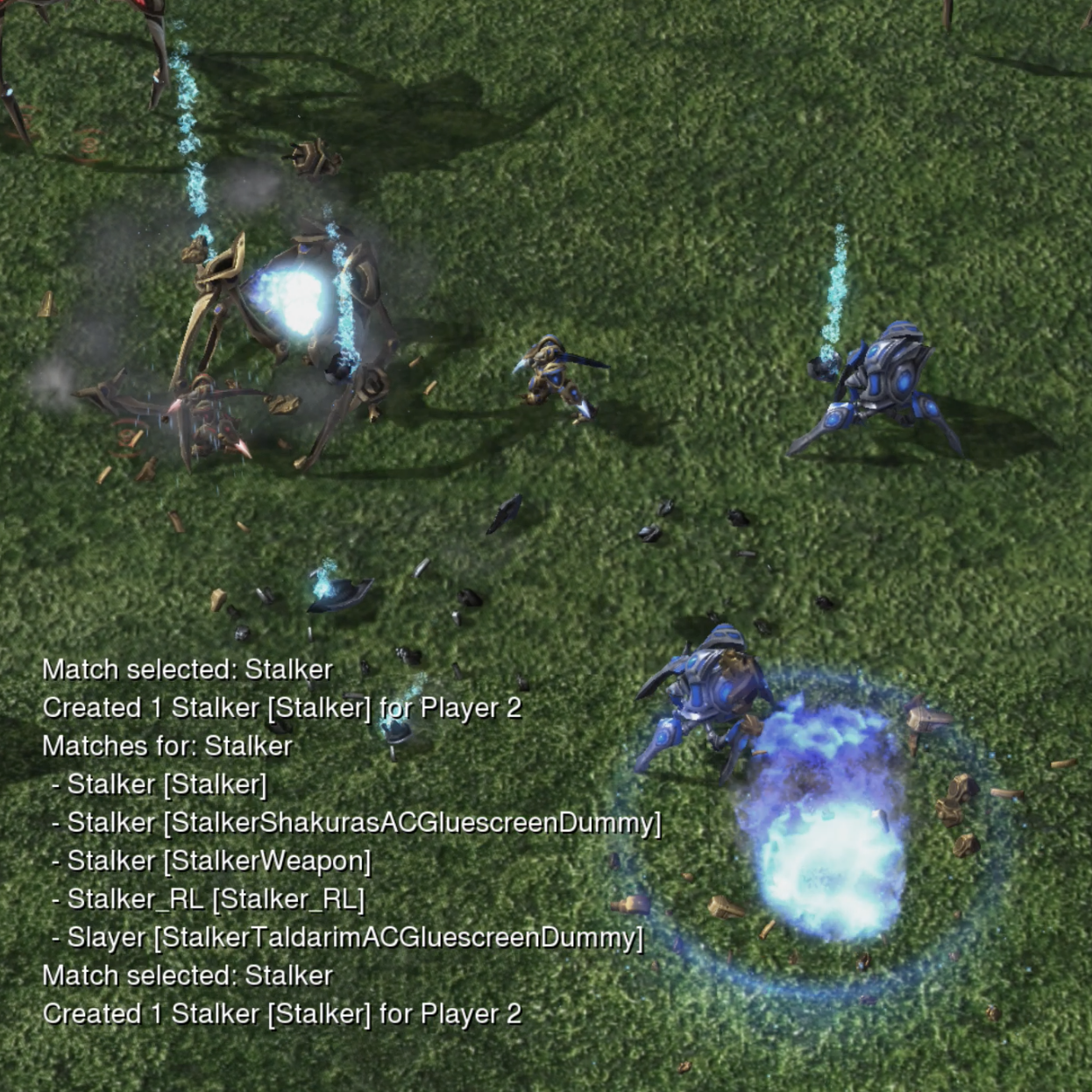}
         \caption{SMACv2}

     \end{subfigure}%
     \begin{subfigure}{\subfigsize\linewidth}
         \centering
         \includegraphics[width=\linewidth,frame]{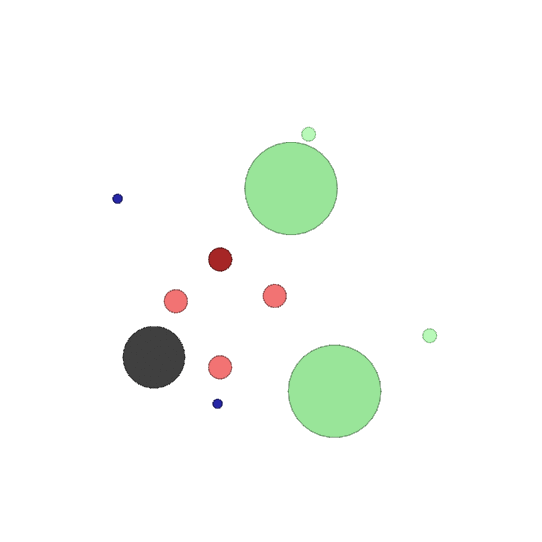}
         \caption{MPE}
    
     \end{subfigure}%
       \begin{subfigure}{\subfigsize\linewidth}
         \centering
         \includegraphics[width=\linewidth,frame]{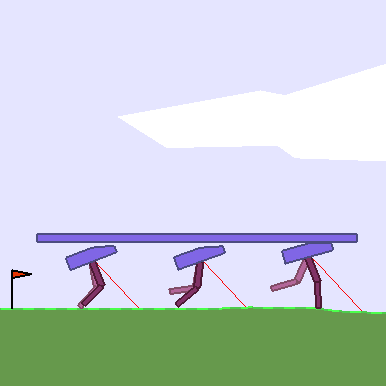}
         \caption{SISL}
      \end{subfigure}%
   \begin{subfigure}{\subfigsize\linewidth}
         \centering
        \includegraphics[width=\linewidth,frame]{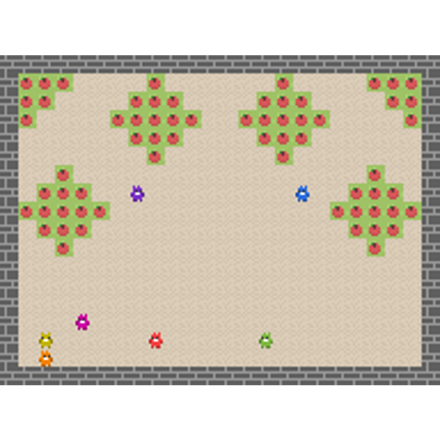}
         \caption{MeltingPot}
     \end{subfigure}
    \caption{Environments in BenchMARL. This figure shows renderings from one example task for each environment. Details and references for all environments are available in \autoref{tab:tasks}.}
    \label{fig:task_renderings}
\end{figure}

\paragraph{Tasks.} Tasks are scenarios from an environment which constitute the MARL challenge to solve. In~\autoref{tab:tasks} we report the environments available in BenchMARL with the respective number of tasks available in the library. \autoref{fig:task_renderings} shows renderings for one example task for each of the environments.
Environments in BenchMARL showcase the variety of MARL paradigms compatible with the library (e.g., differing in cooperation, reward sharing, action space). Importantly, BenchMARL supports vectorized environments, allowing to scale simulation when using batched environments on GPU devices\footnote{For more info, see~\cite{bettini2022vmas}.}. When using a vectorized environment on GPU, both simulation and training can be performed on the same device, granting important performance speed-ups and avoiding data casting and CPU-GPU communication costs. Tasks are represented as enumerations of an environment in order to provide auto-completion functionality when choosing a task and to ground the available tasks associated to a specific BenchMARL release in a static fashion. BenchMARL further supports environments with action masking (e.g., \citep{ellis2022smacv2}) and allows users to specify custom transforms to process the input-output data of an environment. For example, in environments with visual observations such as MeltingPot~\citep{leibo2021meltingpot}, transforms are used to store images as integers in the replay buffer, and appropriately cast them to floating point when passing them to neural networks.

\begin{table}[t]
    \centering
    \caption{Environments in BenchMARL.}
    \label{tab:tasks}
    \resizebox{\linewidth}{!}{%
    \begin{tabular}{lccccccc}
    \hline
        \textbf{Environment} & \textbf{Tasks} & \textbf{Cooperation} & \textbf{Global state} & \textbf{Reward function} & \textbf{Action space} & \textbf{Vectorized}  \\ \hline
        VMAS~\citep{bettini2022vmas} & 27 & Cooperative + Competitive & No & Shared + Independent + Global & Continuous + Discrete & Yes \\ 
        SMACv2~\citep{ellis2022smacv2} & 15 & Cooperative & Yes & Global & Discrete & No \\ 
        MPE~\citep{lowe2017multi} & 8 & Cooperative + Competitive & Yes & Shared + Independent + Global & Continuous + Discrete & No  \\ 
        SISL~\citep{gupta2017cooperative} & 2 & Cooperative & No & Shared & Continuous & No  \\
        MeltingPot~\citep{leibo2021meltingpot} & 49 & Cooperative + Competitive & Yes & Independent & Discrete & No  \\ \hline
    \end{tabular}
    }
\end{table}

\paragraph{Models.}
BenchMARL models are neural network blueprints designed to be used in multiple MARL contexts. They can be instantiated in multiple ways.
(1) \textit{Decentralized}: when a model is instantiated in a decentralized manner (e.g., a policy) it will compute a separate value for each agent. These separate values will only be conditioned on local inputs (e.g., the agent's observation or observations from a local neighborhood).
(2) \textit{Centralized with global input}: when a model is instantiated in centralized mode with global input (e.g., a global critic) it will compute a shared value for a group of agents given a shared input. For example, this could be a centralized critic that takes a global top-down view of the task and outputs a shared value. (3) \textit{Centralized with local input}: like \textit{centralized with global input} with the difference that the value is computed from aggregating local inputs. For example, this could be a centralized critic that takes a concatenation of the agents' observations and outputs a shared value.
Furthermore, in any of these settings, it is possible to decide whether or not to share parameters among agents. Users can decide which model to use for their critics and actors and are able to mix models as they like. The models currently available in BenchMARL are reported in \autoref{tab:models} and include: Multi Layer Perceptron (MLP), Graph Neural Network (GNN), Convolutional Neural Network (CNN), Long Short-Term Memory (LSTM), Gated Recurrent Unit (GRU), and Deep Sets~\citep{zaheer2017deep} models. BenchMARL models can be chained in a sequence, allowing to build more complex networks. For example, by chaining a CNN to a fully-connected GNN attention layer~\citep{velivckovic2018graph} and an MLP, it is possible to process agents' image inputs using a transformer-like architecture.

\begin{table}[t]
    \centering
    \caption{Models in BenchMARL.}
    \label{tab:models}
     \resizebox{\linewidth}{!}{%
     \begin{tabular}{lcccc}
    \hline
        \textbf{Model} & \textbf{Decentralized} & \textbf{Centralized with local inputs	} & \textbf{Centralized with global input}   \\ \hline
         MLP & Yes & Yes & Yes \\
            CNN & Yes & Yes & Yes \\
           GRU & Yes & Yes & Yes \\
            LSTM & Yes & Yes & Yes \\
           GNN & Yes & Yes & No  \\
            Deep Sets~\citep{zaheer2017deep} & Yes & Yes & Yes  \\ \hline
    \end{tabular}}
\end{table}

%% file: sections/features.tex
\label{sec:features}

In the following, we present some of the core features that enable the library's standardization and reproducibility goals. 

\paragraph{Documentation, tests, engineering.}
The library is accompanied by version-tracked documentation available \href{https://benchmarl.readthedocs.io}{here}.
Each component is accompanied by docstrings compiled in the \textit{package reference} section to explain the components' functionalities.
We provide an extensive set of examples and notebooks in the dedicated repository folder with the goal of showcasing the library's main use-cases. Integration and unit tests are run on all tasks and algorithms, with a separate job being run for each environment. These tests are executed in the continuous integration (CI) and perform complete training iterations to check all components. Coverage is reported online and is currently around 90\%. The library is maintained by the authors and the community with its backend maintained directly in the TorchRL project.

\paragraph{Reporting.} The library is directly integrated with the reporting tools proposed by \cite{gorsane2022towards}. This allows users to avoid spending time crafting dedicated plots while providing them with state-of-the-art tools for automating this process. To enable this, BenchMARL experiments optionally output results in the JSON format required by the reporting library. Practitioners will be able to use the tools in \url{https://github.com/instadeepai/marl-eval} to process raw  MARL experiment data for downstream use with the tools provided by \cite{agarwal2021deep}. 
BenchMARL is furtherly compatible with all the loggers available in TorchRL (e.g., Wandb~\citep{wandb}, tensorboard~\citep{tensorflow2015-whitepaper}, csv), with additional support for saving and restoring experiments. Wandb users will be able to access online interactive plots for their experiments together with renderings of evaluation runs.

\paragraph{Configuring.}
A core reproducibility challenge resides in sharing experiment configurations.
For this reason, BenchMARL completely decouples configuration parameters from the Python codebase. Configuration files use a standardized YAML format and are available in the \lstinline[columns=fixed]{conf} folder, structured in sub-folders mimicking the hierarchy of the respective Python components.
When an experiment is instantiated, each component configuration is loaded from YAML into a corresponding Python dataclass. This process allows to strongly check types and values of a configuration and fail early if some parameters do not pass this check. Loaded BenchMARL components' dataclasses can then be safely passed around as they only represent a lazy configuration and their respective component is not instantiated until the experiment is run.
To simplify the process of loading configurations from YAML, users can optionally benefit from using Hydra~\citep{Yadan2019Hydra}, a project 
that allows to define modular configuration trees in YAML files that can be overridden in many ways either within scripts or in the command line. Such modularity in the configuration allows to run complex benchmarks in one line by listing the desired algorithms, models, and tasks to compare. Different execution backends can be used (e.g., sequential, parallel, SLURM). Users that do not want to depend on Hydra can directly load YAML configurations into the respective dataclass from a Python script.

\paragraph{Extending.} Each component in the library has an associated abstract class which defines the minimal functionalities needed to implement a new instance. This makes it easy to integrate custom algorithms, models, and tasks allowing to compare them against the wide range of already implemented ones. Our examples provide detailed illustrations on how to create custom components. For example, when creating a new algorithm, users will just need to implement the functions that provide a TorchRL loss and policy as well as some general flags on whether the algorithm should be trained on or off policy and what type of actions are supported. Similarly, when integrating a new environment, they will need to provide a TorchRL environment class alongside basic information on the input and output spaces, global state, agent grouping, and action type. BenchMARL will then automatically be able to determine compatibility between environments and algorithms.

\paragraph{Callbacks and checkpointing.} To aid in experimentation, BenchMARL provides multiple callbacks in all the major phases of the training process. These callbacks will notify a list of listener components defined by the user. For example, callbacks can be used: to stop or resume training a particular agent group, to train additional losses on top of the RL objective, to create a curriculum of tasks, to perform self-play in adversarial environments, or simply to log custom data or plots. Alongside callbacks, BenchMARL is compatible with experiment checkpointing and restoring in the standard PyTorch format. Experiment state (e.g., replay buffers, networks, environments) can all be loaded and saved to disk. This particular feature also enables the deployment of trained policies, as shown in \cite{blumenkamp2024cambridge}, where policies trained in BenchMARL and VMAS are deployed on a real-world multi-robot system.

\paragraph{Public benchmark results.}
As part of the effort for the standardization of MARL benchmarking, we are fine-tuning and releasing hyperparameters and experiment results for BenchMARL environments and algorithms in public interactive notebooks.
Fine-tuned configurations are available in the \lstinline[columns=fixed]{conf} folder of the project repository.
The notebooks and interactive plots are publicly available on Wandb (at \href{https://wandb.ai/matteobettini/benchmarl-public/reportlist}{this url}).
Towards this goal, we have already run and published benchmarks for VMAS environments. The results are reported in \autoref{sec:experiments}.


%% file: sections/experiments.tex
\label{sec:experiments}

In this section, we report some experiments to confirm the correctness of the implementations in the library and provide public benchmarking references. Furthermore, since the majority of the algorithms available is simply a multi-agent extension of single-agent algorithms (e.g., PPO, DDPG, SAC, DQN), BenchMARL is able to reuse the algorithm implementations available in TorchRL. We refer to \cite{bou2023torchrl} for extended evaluations of these single-agent algorithms as well as comparisons to the original papers.

\begin{figure}[t]
    \centering
    \begin{subfigure}{0.55\linewidth}
    \includegraphics[width=\linewidth]{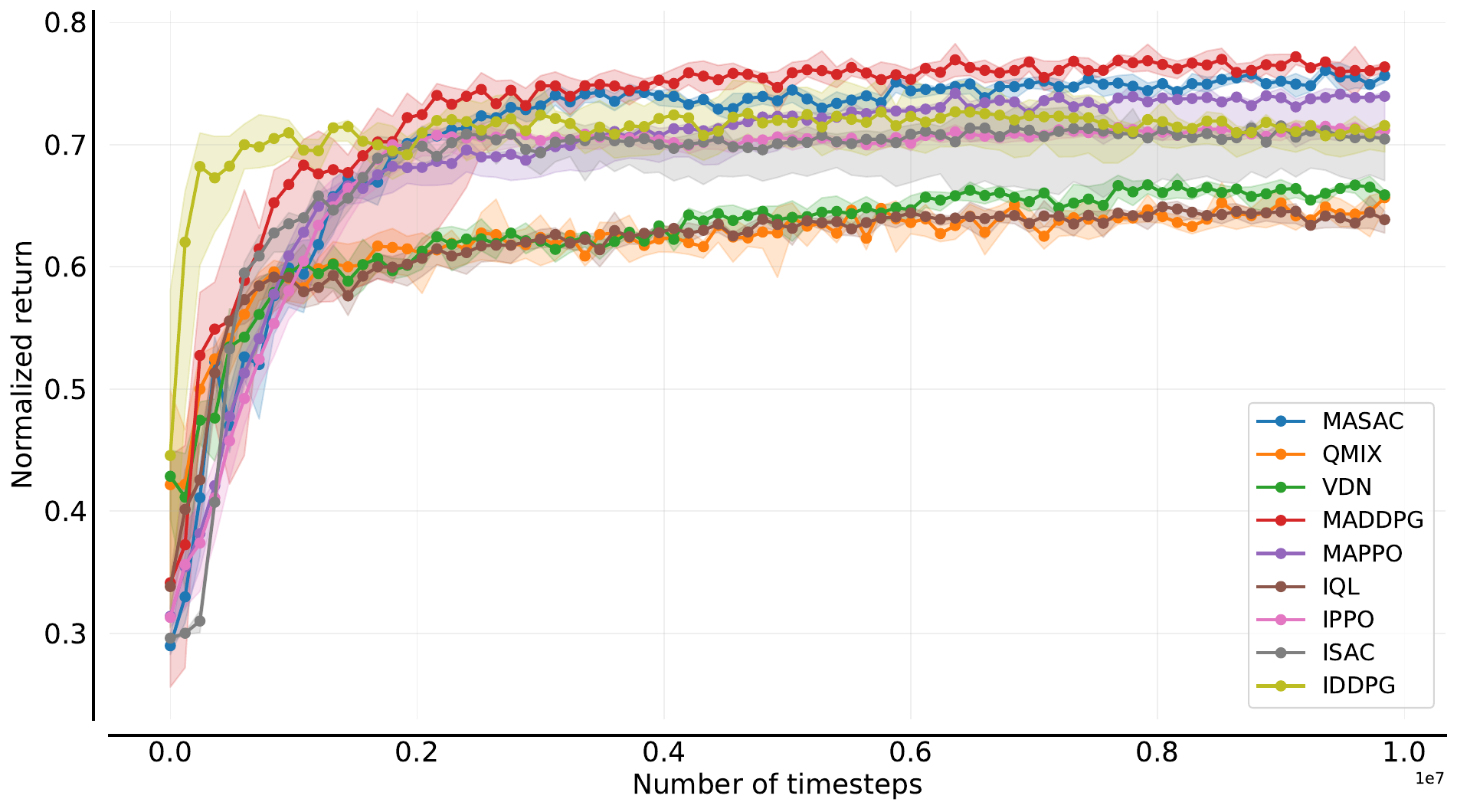}
    \caption{Sample efficiency curves.}
    \label{fig:navigation}
    \end{subfigure}
    \begin{subfigure}{0.42\linewidth}
    \includegraphics[width=\linewidth]{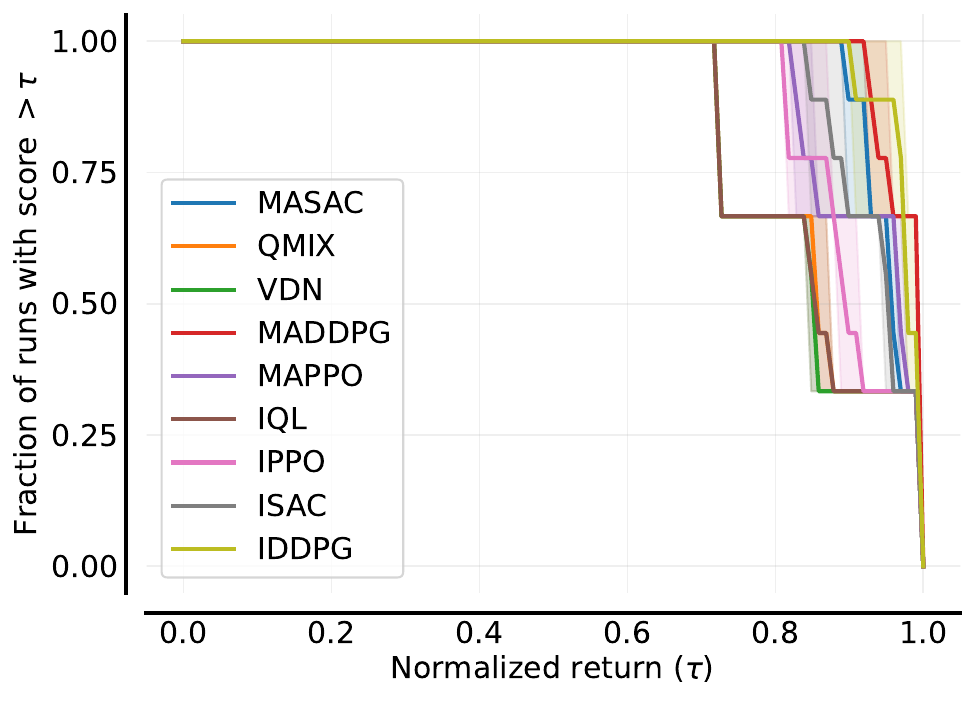}    
    \caption{Performance profile.}
    \label{fig:balance}
    \end{subfigure}
    \begin{subfigure}{\linewidth}
    \includegraphics[width=\linewidth]{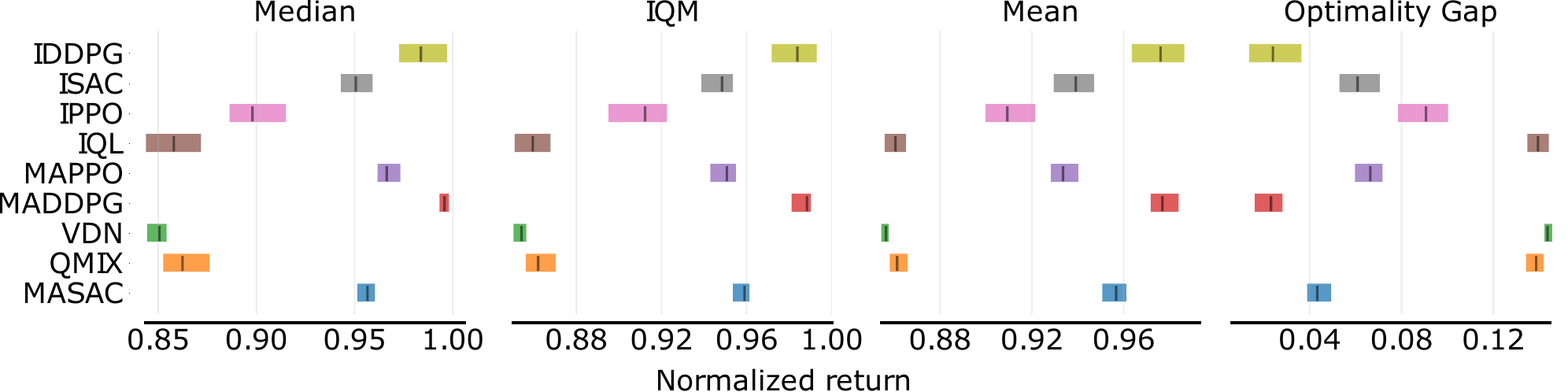}
    \caption{Aggregate scores.}
     \label{fig:sampling}
    \end{subfigure}\hfill
    \caption{Benchmark results over VMAS tasks (\textit{Navigation}, \textit{Sampling}, \textit{Balance}). All plots report the 95\% stratified bootstrap confidence intervals over 3 random seeds for each experiment. Curves in the top report the inter-quartile mean (IQM). See  \citep{gorsane2022towards,agarwal2021deep} for more details on the reported metrics. Details and references for the algorithms used are available in \autoref{tab:algorithms}.}
    \label{fig:experiment_recap}
\end{figure}

\begin{figure}[!ht]
    \centering
    \begin{subfigure}{0.3\linewidth}
    \includegraphics[width=\linewidth]{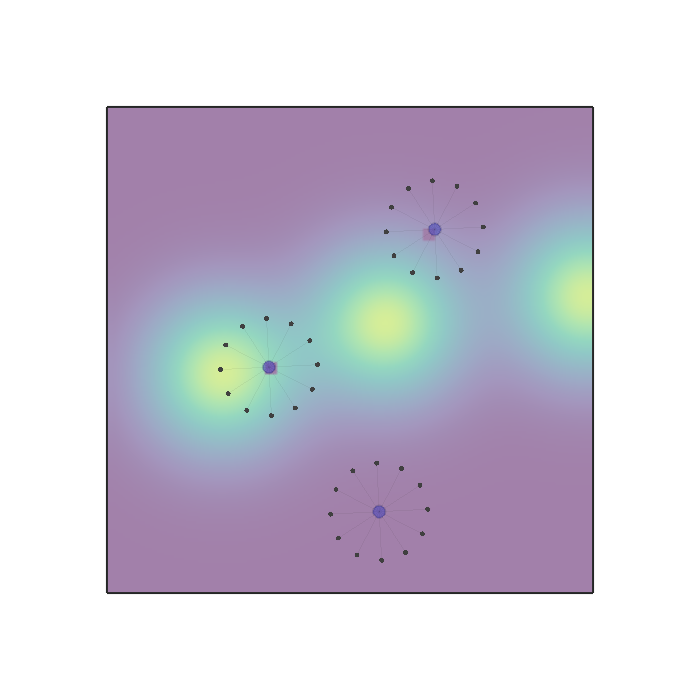}
    \caption{\textit{Sampling} task.}
    \end{subfigure}
    \begin{subfigure}{0.67\linewidth}
    \includegraphics[width=\linewidth]{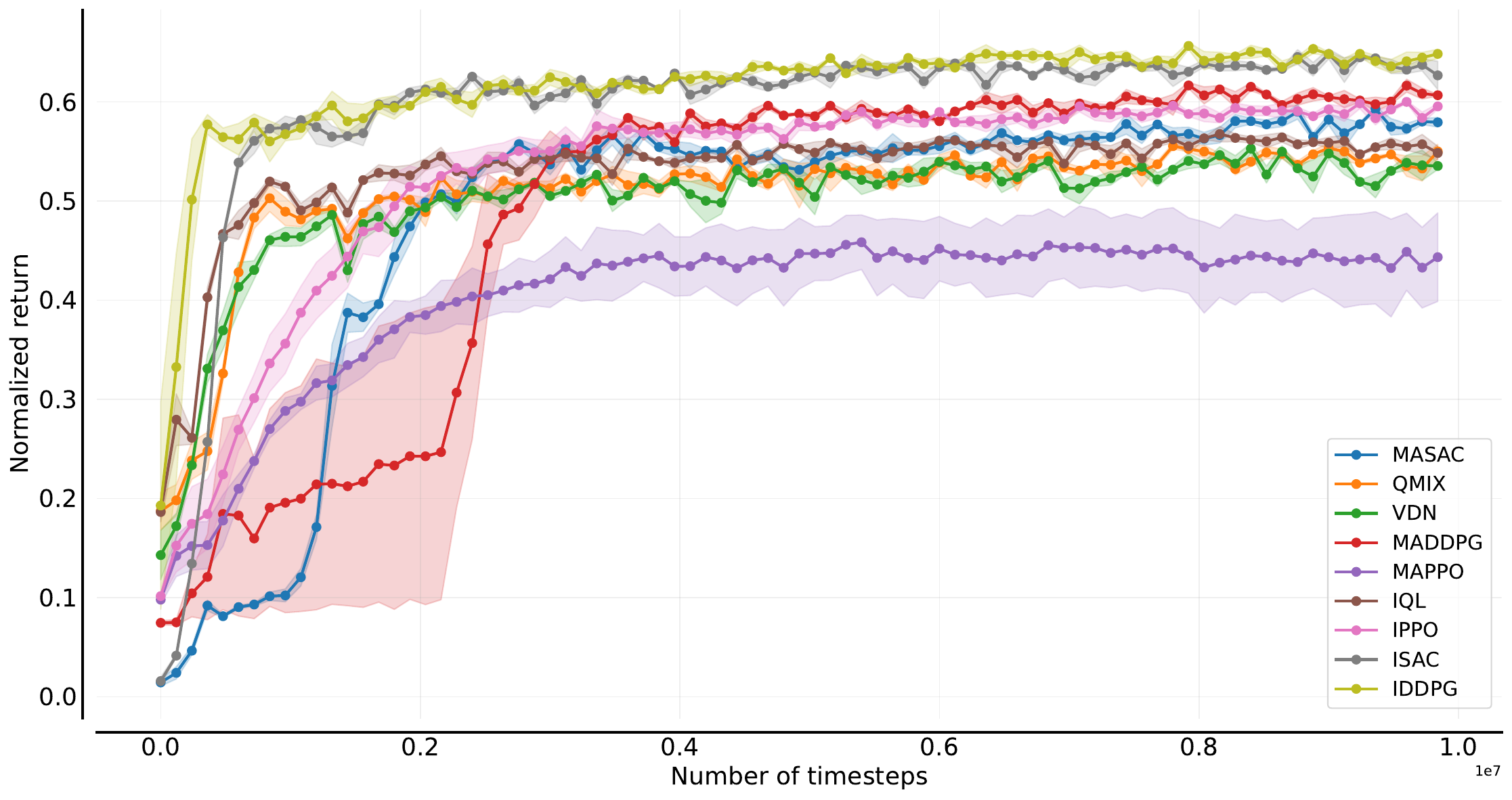}
    \caption{\textit{Sampling} reward curves.}
    \end{subfigure}

    \begin{subfigure}{0.3\linewidth}
    \includegraphics[width=\linewidth]{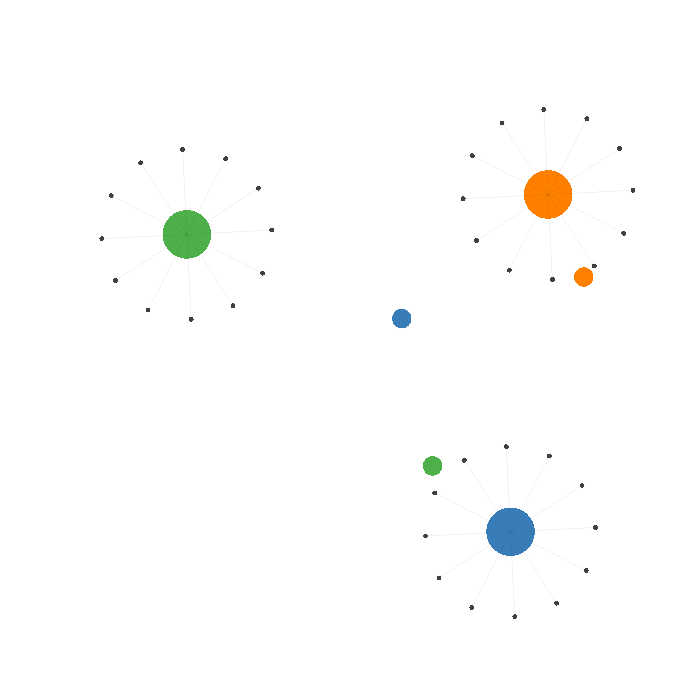}    
    \caption{\textit{Navigation} task.}
    \end{subfigure}
    \begin{subfigure}{0.67\linewidth}
    \includegraphics[width=\linewidth]{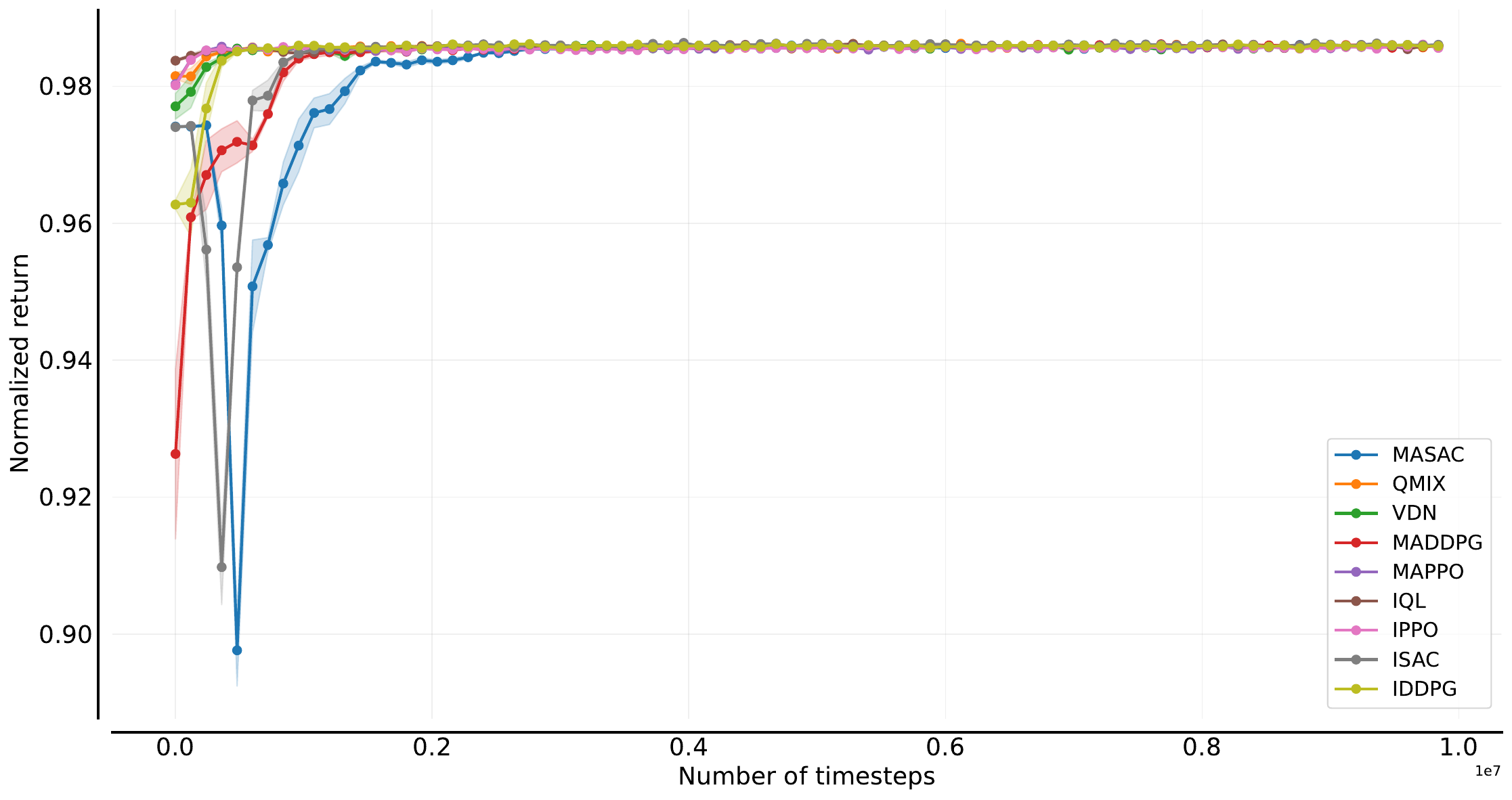}    
    \caption{\textit{Navigation} reward curves.}
    \end{subfigure}

    \begin{subfigure}{0.3\linewidth}
    \includegraphics[width=\linewidth]{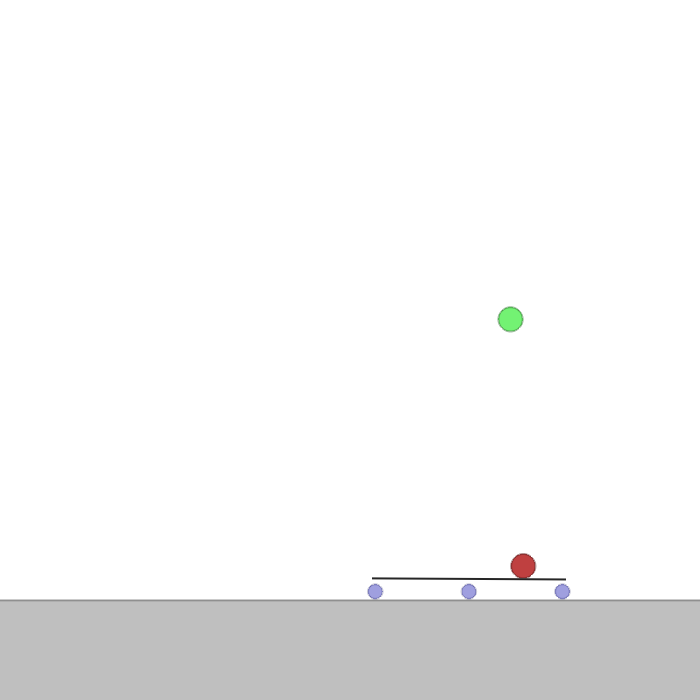}
    \caption{\textit{Balance} task.}
    \end{subfigure}
    \begin{subfigure}{0.67\linewidth}
    \includegraphics[width=\linewidth]{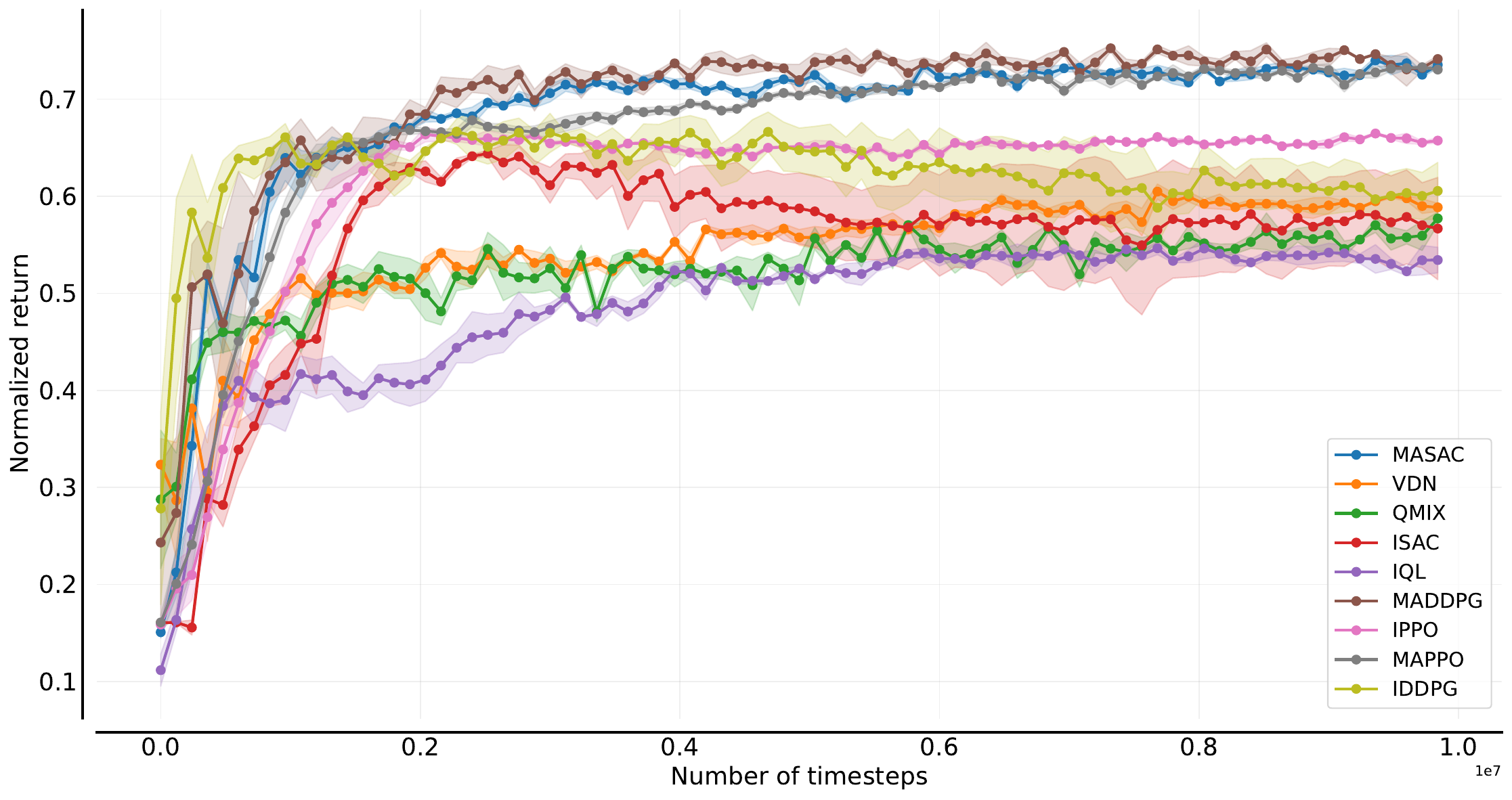}
    \caption{\textit{Balance} reward curves.}
    \end{subfigure}
    \caption{The sample efficiency curves for all BenchMARL algorithms over the three VMAS tasks analyzed. We report the inter-quartile mean (IQM) with 95\% stratified bootstrap confidence intervals over 3 random seeds for each experiment. Details and references for the algorithms used are available in \autoref{tab:algorithms}.}
    \label{fig:individualt_tasks}
\end{figure}

A first set of benchmarks was run on the VMAS tasks.
For these experiments, we ran all of the currently available algorithms in BenchMARL on the \textit{Navigation}, \textit{Sampling}, and \textit{Balance} tasks.
Experiment results, aggregated over all tasks, are reported in \autoref{fig:experiment_recap}. 
Individual task results are reported in \autoref{fig:individualt_tasks}.
All the algorithms, models, and tasks were run using the default BenchMARL configuration, available in the  \lstinline[columns=fixed]{conf} folder. The experiment hyperparameters are available in the \lstinline[columns=fixed]{fine_tuned/vmas} folder.
The obtained results match the ones reported by~\cite{bou2023torchrl} and~\cite{bettini2022vmas}.
An interactive version
of these results is available \href{https://api.wandb.ai/links/matteobettini/wu6wwp6q}{here}.

As we can see from the general plots in \autoref{fig:experiment_recap} (aggregated over all tasks), the Multi-Agent (MA) versions of all algorithms (i.e., MASAC, MADDPG, MAPPO) obtain the best overall performance. This is because these algorithms benefit from centralized critics during training, which allows the critic to condition on the global state instead of using just local information. Q-Learning algorithms (i.e., IQL, VDN, QMIX) perform suboptimally compared to the actor-critic ones. This might be due to the fact that these approaches are trained on a version of the tasks with discrete actions, which might impede performance given the continuous multi-robot control nature of these tasks.

%% file: sections/conclusion.tex
In this paper we present BenchMARL, the first MARL benchmarking library with the goal of enabling standardization and reproducibility in the field. 
BenchMARL focuses on high-level structuring of configuration and reporting while using low-level benchmarked RL implementations from TorchRL.
Thanks to this, it provides a lightweight and easy-to-use MARL training library that has already proved successful in multi-robot learning and zero-shot deployment in the real world~\citep{blumenkamp2024cambridge}.
The MARL community can take advantage of BenchMARL to easily compare and share MARL components, increasing reproducibility in the field and reducing its costs.
The library also provides an easy-to-use tool for users approaching MARL for the first time.